\documentclass{article}
\usepackage{spconf,amsmath,epsfig}
\usepackage{stfloats}
\usepackage{float}
\usepackage{multirow}
\usepackage{tabularx}
\usepackage{amssymb}
\usepackage{url}
\usepackage{xcolor} 
\usepackage{hyperref}
\usepackage{comment}
\usepackage{booktabs}
\usepackage[hang,small,bf]{caption}
\usepackage[subrefformat=parens]{subcaption}
\usepackage{enumitem}

\title{Hyperspectral Image Dataset for Individual Penguin Identification}

\twoauthors
  {Youta Noboru,~~~~~~~~~~~ Yuko Ozasa}
      {Tokyo Denki University\\School of System Design and Technology\\5 Senju Asahi-cho, Adachi-ku, Tokyo}
  {Masayuki Tanaka}
	{Tokyo Institute of Technology\\
	School of Engineering, \\
	2-12-1 Ookayama, Meguro-ku, Tokyo}

\begin{document}
%
\maketitle

\begin{figure*}[b]
	\centering
	\includegraphics[width=0.9\textwidth]{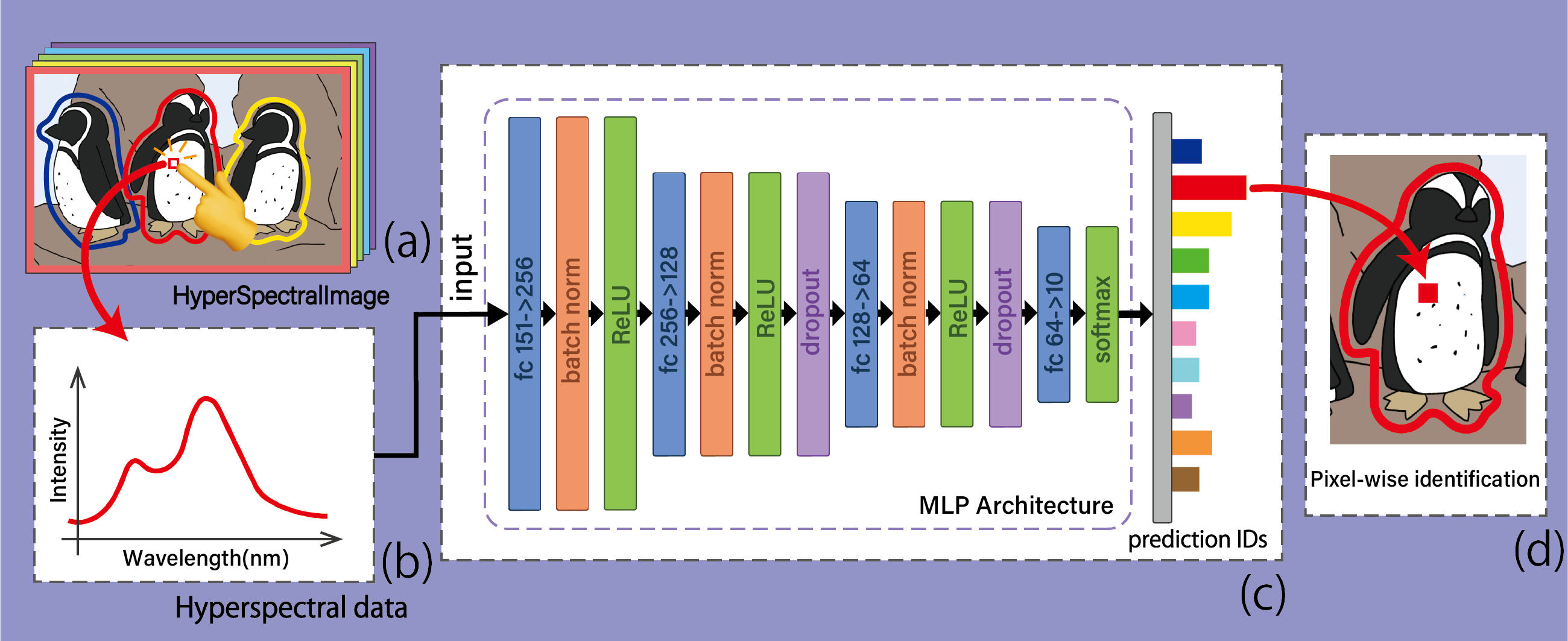}
	\caption{\textbf{Main idea.} (a) Pixels of the target penguins are selected from the captured hyperspectral (HS) images. (b) Pixels selected from the HS images retain information on the wavelength of reflected light. (c) A simple 5-layer MLP is used to predict individual labels of penguins. (d) Individual identification was performed using a single pixel.}
	\label{fig:main}
\end{figure*}

\begin{abstract}
Remote individual animal identification is important for food safety, sport, and animal conservation.
Numerous existing remote individual animal identification studies have focused on RGB images.
In this paper, we tackle individual penguin identification using hyperspectral (HS) images. 
To the best of our knowledge, it is the first work to analyze spectral differences between penguin individuals using an HS camera.
We have constructed a novel penguin HS image dataset, including 990 hyperspectral images of $27$ penguins.
We experimentally demonstrate that the spectral information of HS image pixels can be used for individual penguin identification.
The experimental results show the effectiveness of using HS images for individual penguin identification. The dataset and source code are available here: \textcolor{pink}{\url{https://033labcodes.github.io/igrass24_penguin/}}
\end{abstract}


\section{Introduction}
\label{sec:intro}
Remote individual animal identification is an important task that allows researchers to understand animal behavior and study population parameters such as population size, movement patterns, etc. 
There are invasive and non-invasive methods for individual animal identification.
Invasive methods require capturing an animal and physical tagging, which is costly and likely to cause stress. 
Non-invasive methods use biometric traits such as DNA collected from hair or feces or visual assessment based on images captured in inhabited regions. DNA collection and analysis are costly and sometimes infeasible because they require going to dangerous areas to acquire the samples. 
Therefore, visual assessment and animal biometrics using images are highly demanded. 
\par

In this paper, we focus on individual penguin identification based on hyperspectral (HS) images. Existing image-based animal identification methods \cite{5c6441a934894882a033a76228b24eb8} \cite{Matkowski_2019} \cite{doi:10.1126/sciadv.aaw0736} rely on the spatial information of specific parts of the animal, such as the belly pattern of a penguin or the face of a panda. Those existing image-based methods are infeasible to use for the small size of a pixel of the target. Then, we use HS data of the single pixel. We assume an application with HS images as shown in Fig.~\ref{fig:main}. First, a single pixel in the target penguin is selected from the given HS image. Then, the machine learning model classifies penguin individuals. We analyze HS data of a single pixel because the pixel size of the target penguin in remote sensing is often too small to analyze spatial information.
\par
To the best of our knowledge, it is the first work to analyze spectral differences between penguin individuals using an HS camera. 
There are existing studies identifying deep-sea megafauna using HS image \cite{dumke2018underwater}, however, this work focused on the seafloor and did not capture the individuals.
HS data has also been used in a limited number of studies to differentiate amphibian and reptile species \cite{dodd1981infrared} \cite{pinto2013non} \cite{schwalm1977infrared} and insects \cite{mielewczik2012near}, these studies only analyzed differences between species, not individuals. Kolmann {\it et.al.} distinguishing HS data from $47$ serrasalmid species also documented interspecific variation in pacus that corresponds to cryptic lineages \cite{kolmann2021hyperspectral}.
\par
In this paper, we tackle individual penguin identification using HS data of the single pixel. For this task, we constructed a novel dataset. This dataset comprises $990$ HS images, annotated with $27$ different penguins. Machine learning is used for the identification to evaluate the effectiveness of using HS data. In the experiment, as shown in Fig.~\ref{fig:main}, pixels of the penguins are selected from the HS images, and the selected spectral data are classified into individuals using a machine learning model. 
\par
Our contributions are twofold: First, we examined individual penguin identification using HS data and demonstrated its effectiveness. Second, we constructed a novel dataset for individual penguin identification composed of HS images~\footnote{Our dataset is available here: \textcolor{pink}{\url{https://huggingface.co/datasets/dekkaiinu/hyper_penguin}}}.

\section{Penguin HS image dataset}

\begin{figure}[t]
	\centering
	\includegraphics[width=0.4\textwidth]{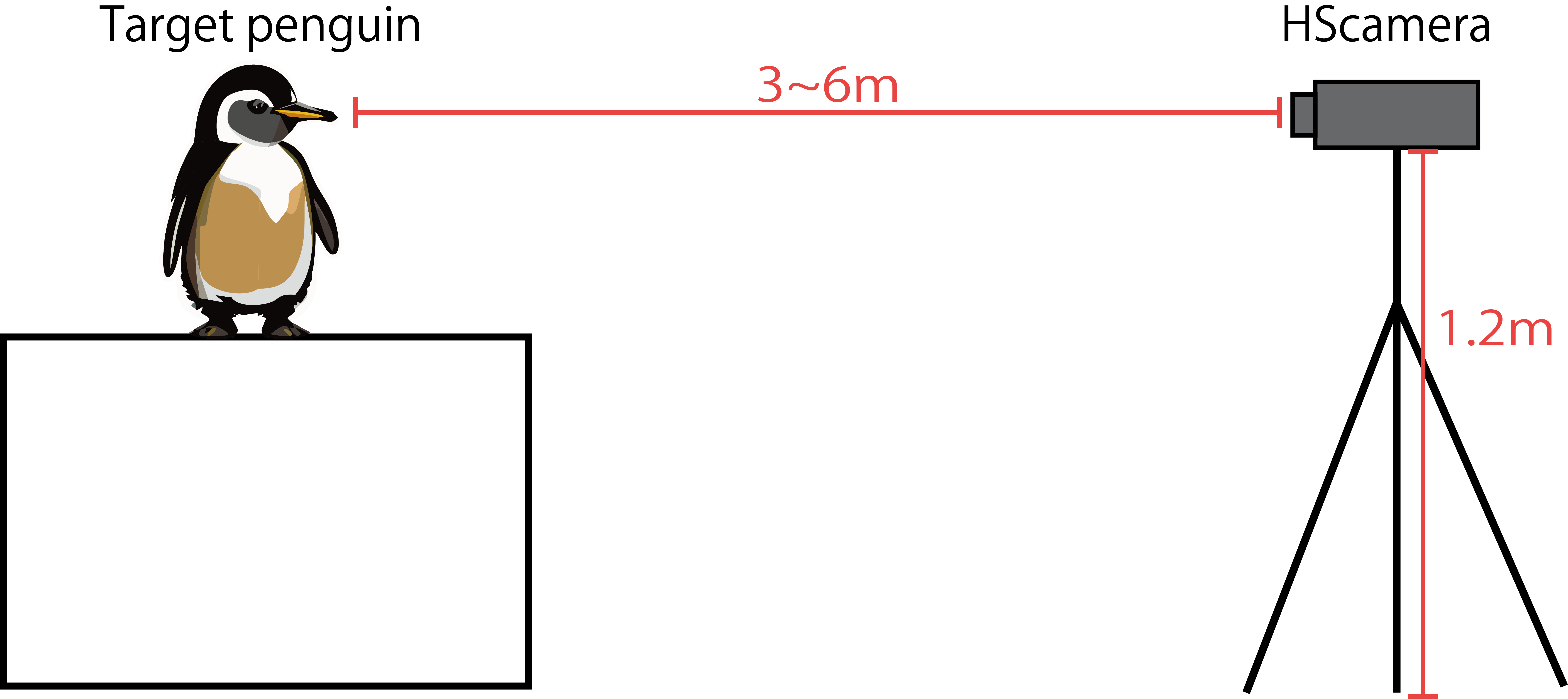}
	\caption{Shooting conditions of the HS images in the dataset.}
	\label{fig:Shooting conditions}
\end{figure}
\begin{figure}[t]
	\centering
	\includegraphics[width=0.4\textwidth]{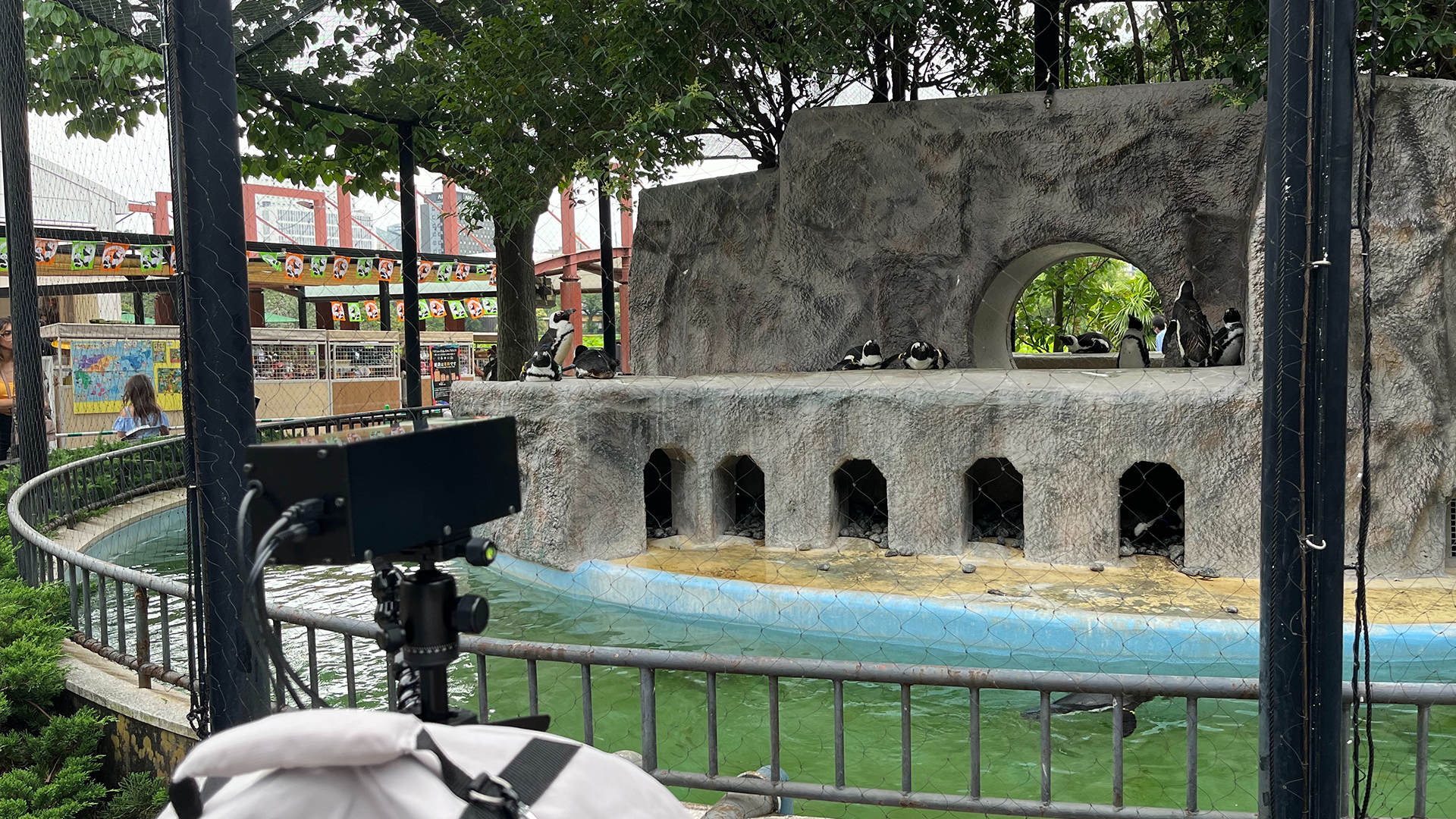}
	\caption{Photographic scenery of the HS images in the dataset.}
	\label{fig:Photographic scenery}
\end{figure}

\begin{figure}[t]
	\centering
	\includegraphics[width=0.4\textwidth]{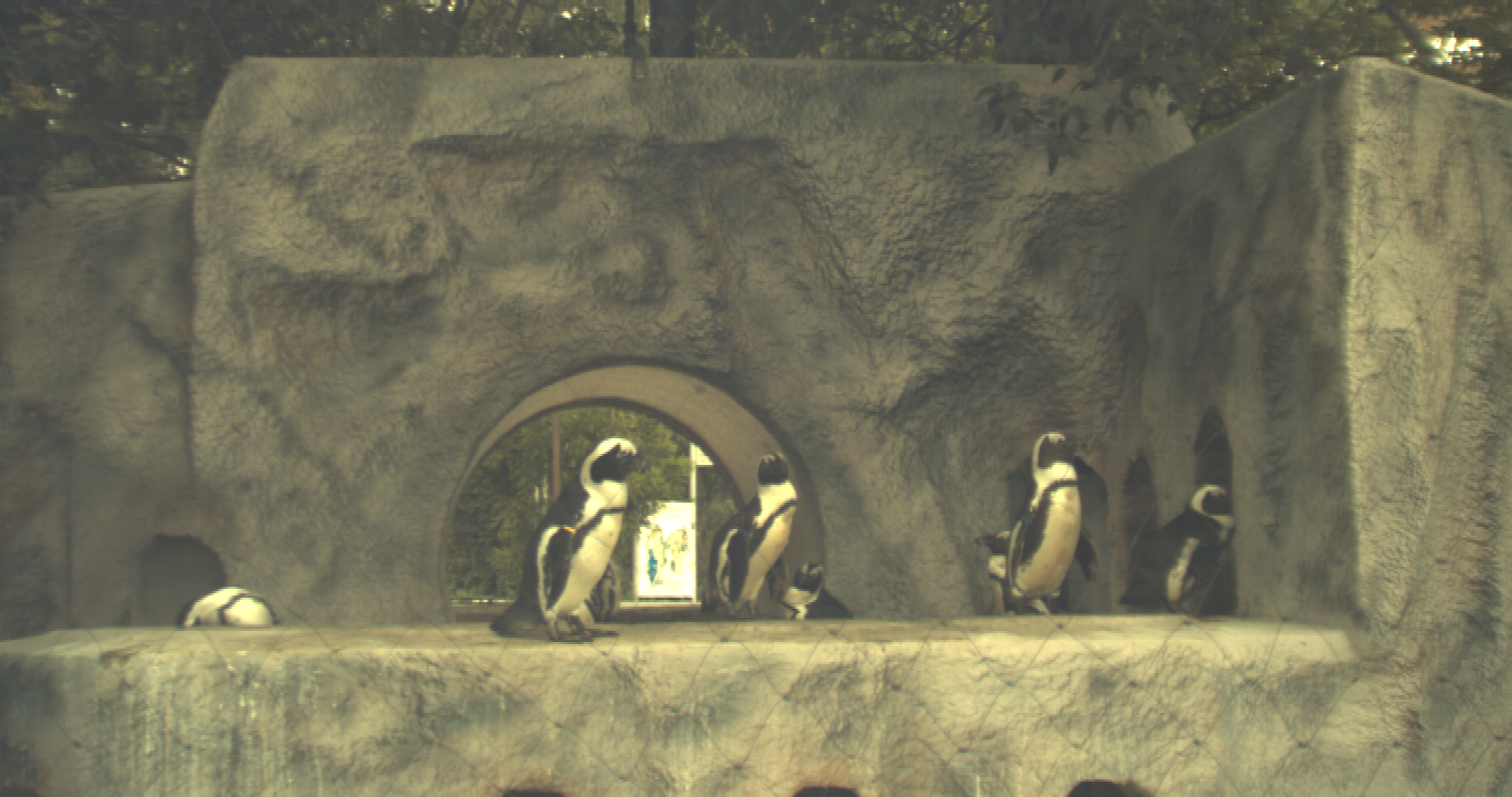}
	\caption{An example from the dataset: An RGB image created from the HS image.}
	\label{fig:Examle Image}
\end{figure}

We collected $990$ HS images of $27$ African penguins at Ueno Zoological Gardens~\cite{UenoZoo}. For collecting HS images, we used the HS camera, which can capture the wave range of $350-1100$[nm], with $151$ bands and a spatial resolution of $5$[nm]~\cite{NH9}. The image size is $2048 \times 1080$ pixels. The distance between the HS camera and the penguins was approximately 3 to 6 meters, with the camera positioned at a height of 1.2 meters, as illustrated in Fig.~\ref{fig:Shooting conditions}. We collected HS penguin images outside, as shown in Fig.~\ref{fig:Photographic scenery}. Additionally, the camera angle was adjusted for each image to ensure that the target penguins were captured in the frame. We captured the HS images in such a manner that a group of penguins is depicted in a single frame.
Each image includes 1 to 6 penguins. Figure~\ref{fig:Examle Image} shows an example of images where an RGB image was converted from an HS image.\par

This dataset is annotated in two ways. The first annotation includes individual IDs of penguins at the pixel level for identifying individuals from (spectral data in) all HS images. The second consists of bounding box annotations marked with individual IDs of penguins, which are used for detecting penguins within the images. Those two ways of annotation allow us to analyze a variety of tasks.

\section{Penguin Identification using HS Images}
We built a learning-based network model for pixel-wise individual penguin identification from the HS data of the single pixel. The network model is a simple $5$-layer multi-layer perceptron (MLP) model with batch normalization, rectified linear unit (ReLU), dropout, and softmax. For the denoising purpose, we applied a $5 \times 5$ sized spatial box filter as preprocessing. Then, we feed the complete information of the single pixel's HS data directly, while the dimension reduction by a principal component analysis (PCA) is recommended in some remote sensing research with HS data~\cite{alhayani2017hyper,uddin2021information}.

\begin{figure*}[t]
	\centering
	\includegraphics[width=0.9\textwidth]{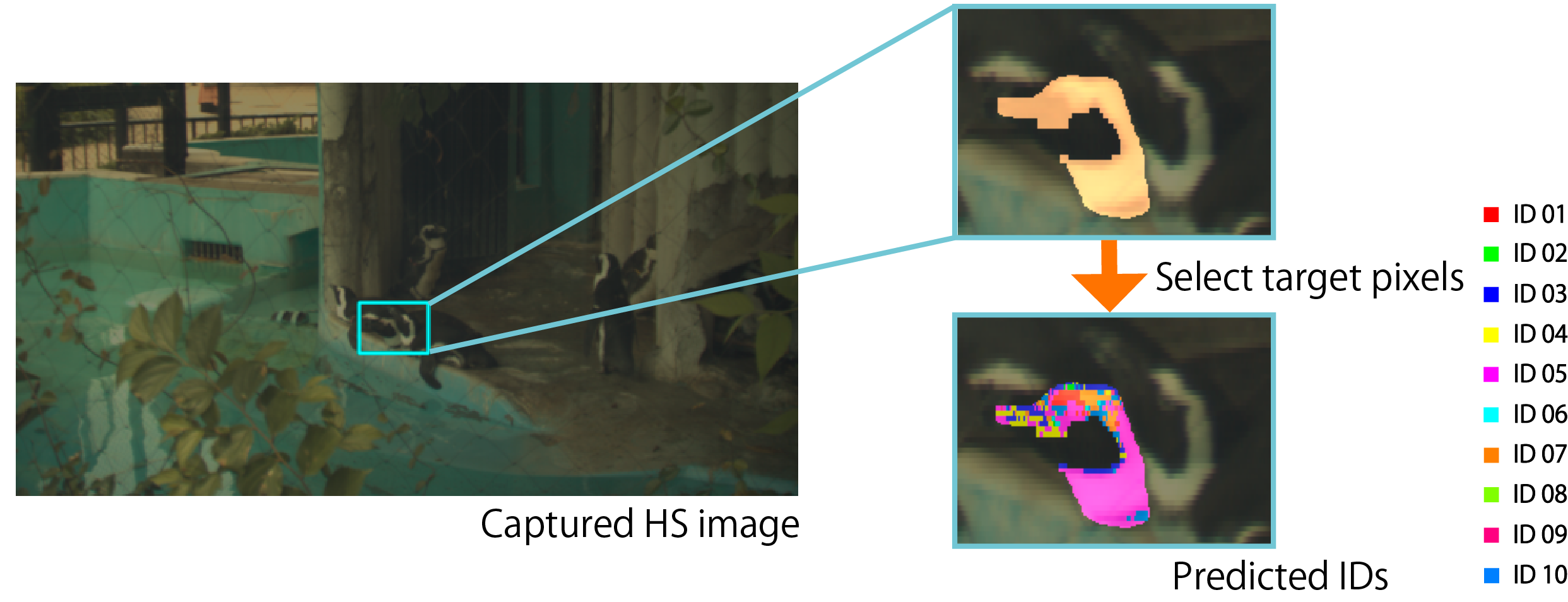}
	\caption{Selection of input pixels with an application in mind and visualization of predicted labels. Pixels of the target with individual label 5 is selected, and the output results are colored differently corresponding to each label.}
	\label{fig:experiments result}
\end{figure*}

\section{Experiments}
We conducted the pixel-wise individual penguin identification based on the HS data. We compare three different input data: RGB data, compressed HS data by the PCA, and the proposed complete HS data. The RGB data was synthesized from the HS data assuming standard RGB color space. For the compressed HS data, we apply the dimension reduction by the PCA. The number of components was set to five, following existing papers~\cite{alhayani2017hyper,uddin2021information}. 

\subsection{Experimental Setup}
\begin{itemize}
    \item 1) Dataset: We picked up ten penguins from our penguin dataset to simplify the problem. Then, annotated HS data were split into the training, the validation, and the test data. 
    For a fair assessment, we selected pixels from different HS images for each dataset. 
    In the training, we used $500,000$ data points, selecting $50,000$ from each penguin id. $50,000$ data points were used for the validation and the test, respectively, where each penguin id has $5,000$ data points.
    \item 2) Evaluation Metrics:
    We quantitatively evaluated the identification performance of each input pixel using Overall Accuracy (OA).
    \item 3) Implementation Details: 
    The experiments we conducted were implemented on the Pytorch platform using a workstation with i9-$10920$X CPU, $128$-GB RAM, and an NVIDIA GeForce RTX $3090$ $24$-GB GPU. We set the number of training epochs to $300$. We adopt the Adam optimizer with a minibatch size of $128$. The learning rate is initialized with $1 \times 10^{-3}$ and decayed by multiplying a factor of $0.6$ after each one-tenth of the total epochs. 
\end{itemize}

\subsection{Experimental Results}

\begin{table}[t]
    \centering
    \caption{Classification accuracy for different data types (\%).}
    \label{table:accuracy}
	\scalebox{0.65}{
	\begin{tabular}{c||c c c c c c c c c c|c}
	  \specialrule{1.5pt}{1pt}{1pt}
	  data & \multicolumn{10}{c|}{Penguin ID.} & Ave. \\
  
	  type&1&2&3&4&5&6&7&8&9&10&(\%)\\
	  \hline
	  \hline
	  RGB &22.5&8.5&49.7&32.1&44.74&16.1&28.8&24.4&13.7&31.0&27.16 \\
        PCA &36.7&30.6&67.6&48.9&64.9&43.6&56.0&46.4&60.4&55.2&51.03 \\
	  HS (Pro.) &\bf{70.8}&\bf{68.7}&\bf{83.7}&\bf{73.3}&\bf{94.6}&\bf{82.3}&\bf{86.3}&\bf{79.3}&\bf{93.4}&\bf{88.2}&{\bf 82.06} \\
	  \specialrule{1.5pt}{1pt}{1pt}
	\end{tabular}}
\end{table}

Table~\ref{table:accuracy} summarizes identification accuracy for each data type. From those results, we can find that the proposed full HS analysis archives $82.06$ [\%] in the average accuracy, while the RGB data analysis and the compressed HS data by the PCA have only $27.16$ [\%], and $51.03$ [\%], respectively. The average accuracy of the proposed method may not be perfect, but we think the proposed method can help humans with individual penguin identification tasks. Further improvement of accuracy includes our future work.
\par
\begin{figure*}[t]
  \centering
  \includegraphics[width=0.9\textwidth]{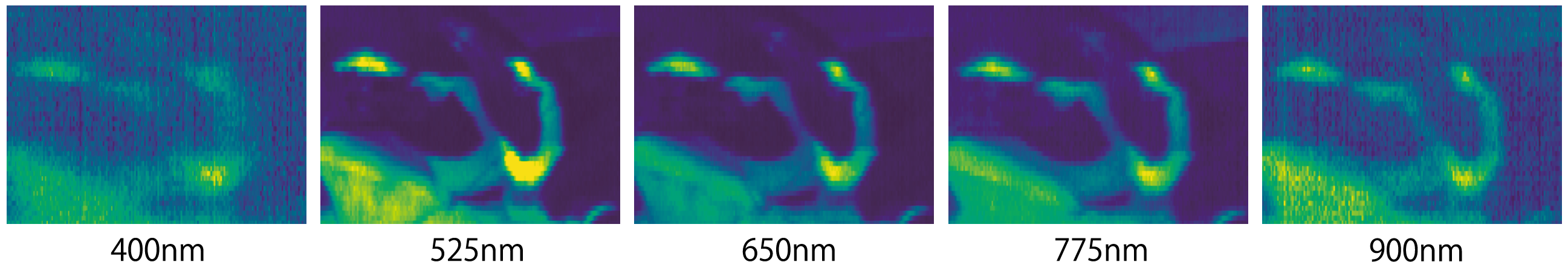}
  \caption{The band images. The band image shows the spectral intensity in each band region. The more yellow the image, the higher the intensity of the corresponding wavelength.}
  \label{fig:experiments result2}
\end{figure*}

We also visualize the pixel-wise identification results. Figure~\ref{fig:experiments result} shows the visualization of the identification results of penguin ID 05, where color represents inference penguin IDs. We can see many pixels are correctly identified, showing a magenta color (ID 05). 
Furthermore, individual identification can be conducted even in cases where only a part of the target penguin is visible due to other penguins appearing in the foreground, provided that the identification is performed on a pixel-wise basis.
The images of Fig.\ref{fig:experiments result2} are band images of the HS image corresponding to the upper right image in Fig.\ref{fig:experiments result}. 
The band images of HS images differ by wavelength, and the HS images contain richer information than those of RGB images while the RGB image is composed of only three band images.

\section{Discussion}

\begin{figure*}[htbp]
  \centering
  \begin{minipage}{.34\textwidth}
    \includegraphics[width=\linewidth]{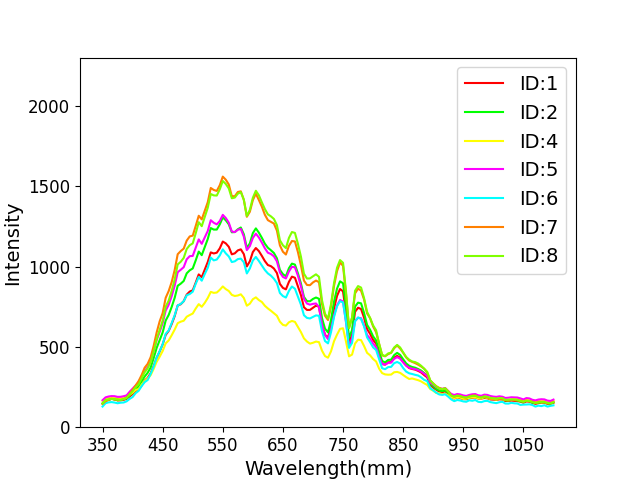}
    \subcaption{June $23$th  10:00-11:00}
    \label{(a)}
  \end{minipage}%
  \begin{minipage}{.34\textwidth}
    \includegraphics[width=\linewidth]{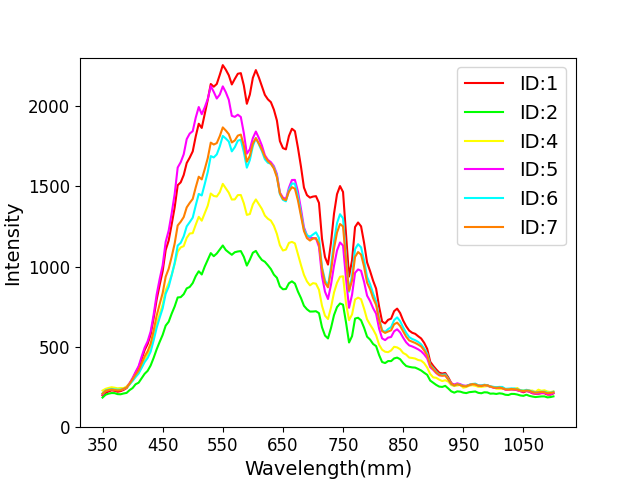}
    \subcaption{June $23$th  12:00-13:00}
    \label{(b)}
  \end{minipage}%
  \begin{minipage}{.34\textwidth}
    \includegraphics[width=\linewidth]{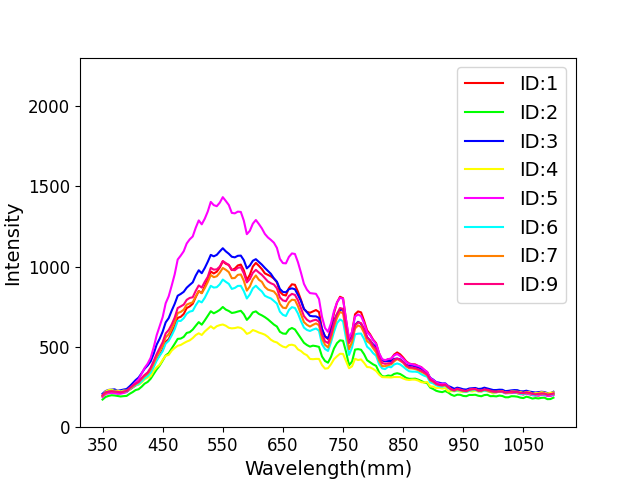}
    \subcaption{June $27$th  13:00-14:00}
    \label{(c)}
  \end{minipage}%
  \caption{Average spectral for each individual at each intervals.}
  \label{fig:spectra}
\end{figure*}

We visualize the differences in spectral among penguin individuals to discuss our experimental results. Figure~\ref{fig:spectra} represents the average spectral of penguins captured in HS images taken over specified one-hour intervals, with each spectral corresponding to a different individual. Figure~\ref{(a)} covers images from $10$ am on June $23$rd for one hour, Figure~\ref{(b)} from $2$ pm on the same day, and Figure~\ref{(c)} from $12$ pm on June $27$th for one hour. The plots depict the average values of the HS data obtained from the white parts of each penguin's body, representing the individual's spectral.
\par
To minimize the influence of sunlight conditions, we visualized the HS data by dividing them into hourly intervals in Figure~\ref{(c)}.
We captured our dataset over one day to obtain the data with various variations since HS data is highly sensitive to sunlight conditions. 
The strongest spectral intensity, evident in Figure~\ref{(c)}, corresponds to the time of day when sunlight is most intense.
\par
The spectral shape of each individual is different for all intervals in  Figure~\ref{fig:spectra}, so the average accuracy was high (shown in Table~\ref{table:accuracy}).
Further, the distinctive spectral of the penguin ID$5$ in Figure~\ref{(b)} and Figure~\ref{(c)} seem to contribute to the highest classification accuracy of $94.6$ [\%]. Thus, the results of the spectral plots demonstrate the effectiveness of HS data in individual identification.

\section{Conclusion}
This paper has presented the individual penguin identification based on HS data of the single pixel. For that purpose, first, we built the penguin HS image dataset with annotation. Then, we experimentally demonstrated that we can identify individual penguins with HS data of the single pixel by the simple MLP network. In future work, we will propose a novel data augmentation method for penguin identification, and we will try to recognize individual animals other than penguins.

\section*{Acknowledgement}
We thank and honor the Ueno Zoological Gardens \cite{UenoZoo} for allowing us to capture HS images of African penguins for the future of image processing research.

\bibliographystyle{IEEEbib}
\bibliography{refs}

\end{document}